# Forward-Forward Training of an Optical Neural Network

Ilker Oguz[1,*], Junjie Ke[2], Qifei Wang[2], Feng Yang[2], Mustafa Yildirim[1], Niyazi Ulas Dinc[1], Jih-Liang Hsieh[1], Christophe Moser[1] and Demetri Psaltis[1]

[1] IEM, EPFL, Switzerland

[2] Google Research, USA

*ilker.oguz@epfl.ch

**Abstract**

Neural networks (NN) have demonstrated remarkable capabilities in various tasks, but their computation-intensive nature demands faster and more energy-efficient hardware implementations. Optics-based platforms, using technologies such as silicon photonics and spatial light modulators, offer promising avenues for achieving this goal. However, training multiple trainable layers in tandem with these physical systems poses challenges, as they are difficult to fully characterize and describe with differentiable functions, hindering the use of error backpropagation algorithm. The recently introduced Forward-Forward Algorithm (FFA) eliminates the need for perfect characterization of the learning system and shows promise for efficient training with large numbers of programmable parameters. The FFA does not require backpropagating an error signal to update the weights, rather the weights are updated by only sending information in one direction. The local loss function for each set of trainable weights enables low-power analog hardware implementations without resorting to metaheuristic algorithms or reinforcement learning. In this paper, we present an experiment utilizing multimode nonlinear wave propagation in an optical fiber demonstrating the feasibility of the FFA approach using an optical system. The results show that incorporating optical transforms in multilayer NN architectures trained with the FFA, can lead to performance improvements, even with a relatively small number of trainable weights. The proposed method offers a new path to the challenge of training optical NNs and provides insights into leveraging physical transformations for enhancing NN performance.

**Main Text**

Neural networks (NN) are among the most powerful algorithms today. By learning from immense databases, these computational architectures can accomplish a wide variety of sophisticated tasks [1]. These tasks include understanding languages, translating between them [2], creating realistic images from verbal prompts [3], and estimating protein structures from genetic code [4]. Given the significant potential impact of NNs on various areas, the computation-intensive nature of this algorithm necessitates faster and more energy-efficient hardware implementations for NNs to become ubiquitous.

With its intrinsic parallelism, high number of degrees of freedom and low-loss information transfer capability, optics offer different approaches for the realization of a new generation of NN hardware. Silicon photonics based modulator meshes have been demonstrated capable of performing tasks that forms the building blocks of NNs, such as linear matrix operations [5] or pointwise nonlinear functions [6]. In addition to the chip-based platforms, two-dimensional spatial light modulators can exploit optics' 3D scalability fully as free-space propagation provides connectivity between each location on the modulator [7,8]. Another approach, reservoir computing, capitalizes on the complex interactions of various optical phenomena to make inferences by training a single readout layer to map the state of the physical system [9–11]. However, in order to achieve the state-of-the-art performance in sophisticated tasks, training of multiple trainable layers is generally required.

Within the framework of NNs, training parameters before a physical layer constitutes a challenge because complex physical systems are difficult to characterize or to analytically describe. Without a fully known and differentiable function to represent the optical system, the error backpropagation (EBP) algorithm which trains most conventional NNs, cannot be used. One solution to this problem utilizes a different NN in the digital domain (a digital twin) to model the optical system. Error gradients of layers preceding the physical system are approximated with the digital twin [12]. However, this method requires a separate experiment characterization phase before training the main NN, which introduces a computational overhead that may be substantial depending on the complexity of the physical system. Another approach resorts to metaheuristic methods by only observing the dependency of the training performance on the values of programmable weights, without modeling the input-output relation of the physical system [13]. The computational complexity of the training is much smaller in this case but the method does not scale well for a large number of parameters.

The Forward-Forward Algorithm (FFA) defines a local loss function for each set of trainable weights, thereby eliminating the need for EBP and perfect characterization of the learning system while scaling efficiently to large numbers of programmable parameters [14]. With this approach the error at the output of the network does not need to be backpropagated to every layer. The local loss function is defined as the goodness metric $L_{goodness}(y) = \sigma(\sum_j y_j^2 - \theta)$, where $\sigma(x)$ is the sigmoid nonlinearity function, $y_j$ is the activation of the $j$-th neuron for a given sample and $\theta$ is the threshold level of the metric. The goal for each trainable layer is to increase L-goodness for positive samples and decrease it for negative samples. For a classification problem, such as the MNIST-digits dataset, the positive samples are created by designating an area within the input pattern for encoding the class of the image. Similarly, a negative sample is created by marking the designated area with a different pattern corresponding to the second class. The difference between the squared sum of activations of positive and negative samples is balanced between each trainable layer with a normalization step to avoid that each layer learns only one of the two representations.

The local FFA training scheme enables the use of analog hardware for NNs [14,15], because, unlike EBP, FFA does not require direct access to or modeling of the weights of each layer in the NN. In our study, we explore this potential of the FFA and experimentally demonstrate that a complex nonlinear optical transform such as nonlinear propagation in a MMF can be incorporated into a NN to improve its performance.

The optical apparatus we used to implement the network which we trained with the FFA is shown in Figure 1. We use multimode nonlinear wave propagation of spatially modulated laser pulses in an optical fiber. Even though the proposed training method is suitable for virtually any system capable of high-dimensional nonlinear interactions, this experiment is selected as the demonstration setup due to its remarkable ability to provide these effects with very low power consumption (6.3 mW average power, 50 nJ per pulse) [16]. The propagation of 10 ps long mode-locked laser (Amplitude Laser, Satsuma) pulses with 1030 nm wavelength in a confined area (diameter of 50 μm) for a long distance (5 m) provides nonlinear interactions between 240 spatial eigenchannels of the MMF (OFS, bend-insensitive OM2, 0.20 NA) using only 50 nJ pulse energy.

Before coupling light pulses to the MMF, their spatial phase is modulated with input data by a phase-only two-dimensional spatial light modulator (Meadowlark HSP1920-600-1300). The input laser beam, approximated as a Gaussian profile $E_{input}(x,y) = E_0 \exp\left(-\frac{(x^2+y^2)}{w_0^2}\right)$, is phase modulated by the SLM. The modulated light beam can be written as $E_{modulated}(x,y) = E_0 \exp\left(-\frac{(x^2+y^2)}{w_0^2}\right) \exp(i\,D(x,y))$, where $D(x,y)$ is the data transferred from the digital domain to optical system, mapped to the range 0 to $2\pi$, $w_0$ is the beam waist size, and $E_0$ is input field. The modulated beam is coupled to the MMF with a plano-convex lens. The output of the MMF is collimated with a lens and its diffraction off a dispersion grating (Thorlabs GR25-0610) is recorded with the camera (FLIR BFS-U3-31S4M-C). As the diffraction angle has a dependency on the wavelength, the dispersion grating enables the camera to capture information about the spectral changes in addition to the spatial changes due to the nonlinearities inside the MMF.

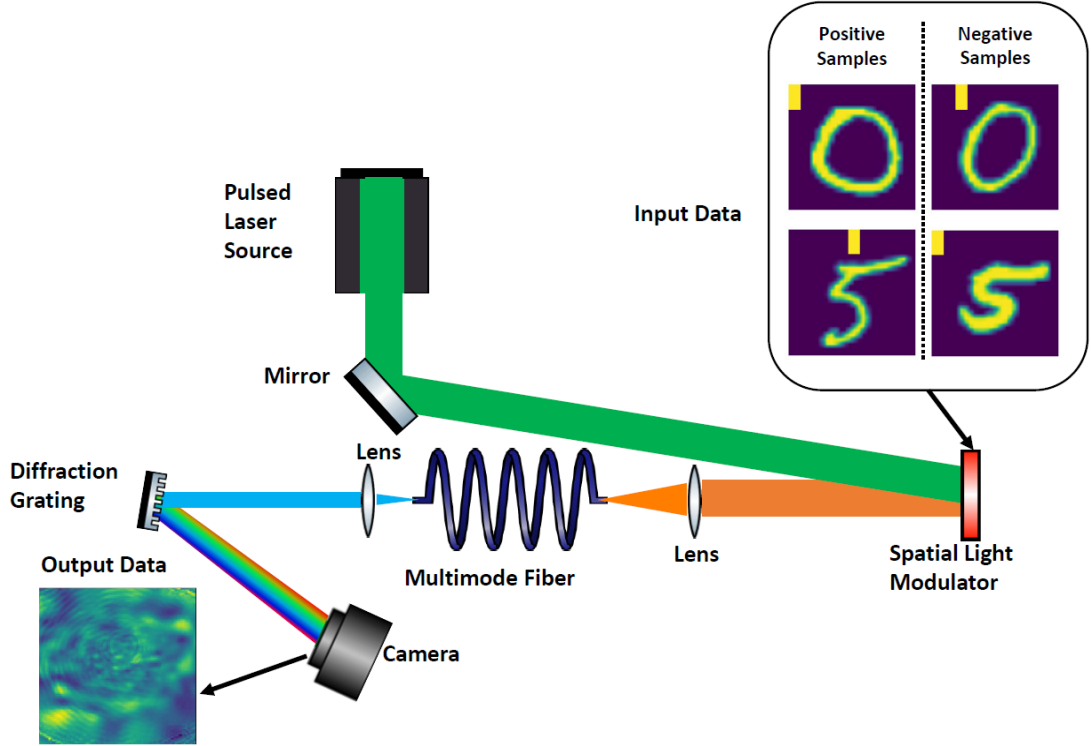

Fig. 1. The schematic of the experimental setup used for obtaining nonlinear optical information transform.

The linear and nonlinear optical interactions in the MMF can be simplified as follows by the multi-modal nonlinear Schrodinger's Equation in terms of the coefficients of propagation modes ($A_p$) of the MMF:

$$\frac{\partial A_p}{\partial z} = \underbrace{i\delta\beta_0^p A_p - \delta\beta_1^p \frac{\partial A_p}{\partial t} - i\frac{\beta_2^p}{2}\frac{\partial^2 A_p}{\partial t^2}}_{\text{Dispersion}} + \underbrace{i\sum_n C_{p,n} A_n}_{\text{Linear mode coupling}} + \underbrace{i\frac{n_2 \omega_0}{A}\sum_{l,m,n} \eta_{p,l,m,n} A_l A_m A_n^*}_{\text{Nonlinear mode coupling}}$$

where $\beta_n$ is the n-th order propagation constant, C is the linear coupling matrix, $n_2$ is the nonlinearity coefficient of the core material, $\omega_0$ is the center angular frequency, A is the core area and $\eta$ is the nonlinear coupling tensor.

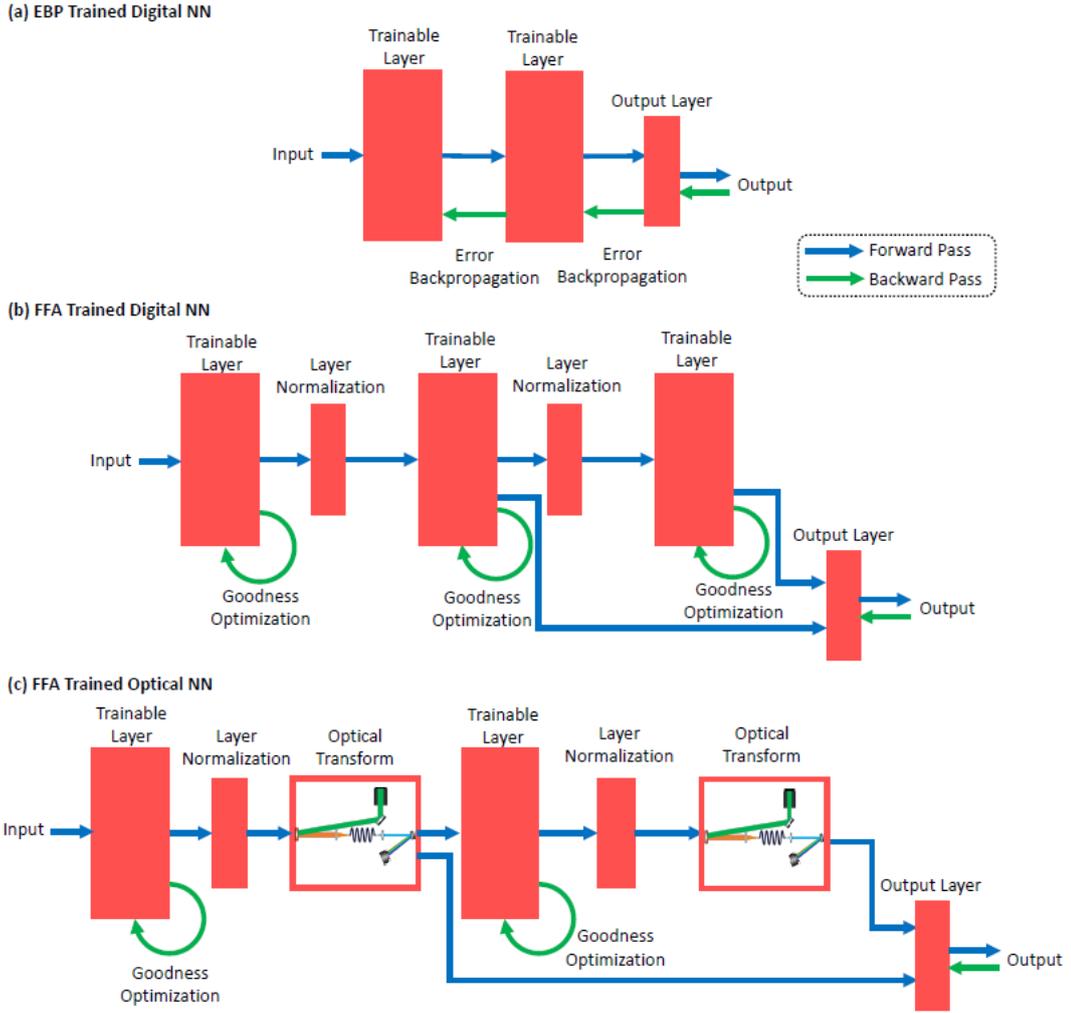

Fig. 2. Different NN architectures compared in the study. (a) A conventional NN trained with error backpropagation. Blue arrows show the information flow in the forward (inference) mode and green lines indicate the training. (b) Diagram of a fully digital NN trained with the FF algorithm. Layers are trained locally with the goodness function. The activations of trainable layers except the first one is used by a separate output layer. (c) Our proposed method also includes optical information transformations between each trainable block. The activations reach to the output layer after the optical transformations.

This equation delineates the nature of interactions obtained with the proposed experiment. In addition to linear coupling, the nonlinear coupling is provided with the multiplication of three different mode coefficients, demonstrating the high-dimensional complexity of the optical interactions.

We evaluated the effectiveness of the proposed approach by constructing a network to implement the MNIST handwritten digits classification task [17]. Due to speed and memory limitations, we randomly selected 4000 samples from the dataset for training, while the validation and test sets were allocated 1000 samples each. The architecture of our neural network is shown in Fig.2. Figure 2a shows a fully digital implementation of a multi-layer network trained with EBP while Figure 2b shows a fully digital implementation trained with the FFA. Finally, Figure 2c includes the optical layers which are trained with the FFA. In all three cases, each layer has similar numbers of trainable parameters and are trained on the same subset of samples with 32x32 resolution. In our implementation, all three NNs start with convolutional layers (2 or 3) followed by a fully connected (FC) output layer of 10 neurons. In the networks shown in Fig. 2.b and 2.c, the output layer is trained with the Ridge classifier algorithm from the scikit-learn library in Python, since this algorithm allows for a faster training with a single-step of singular value decomposition.

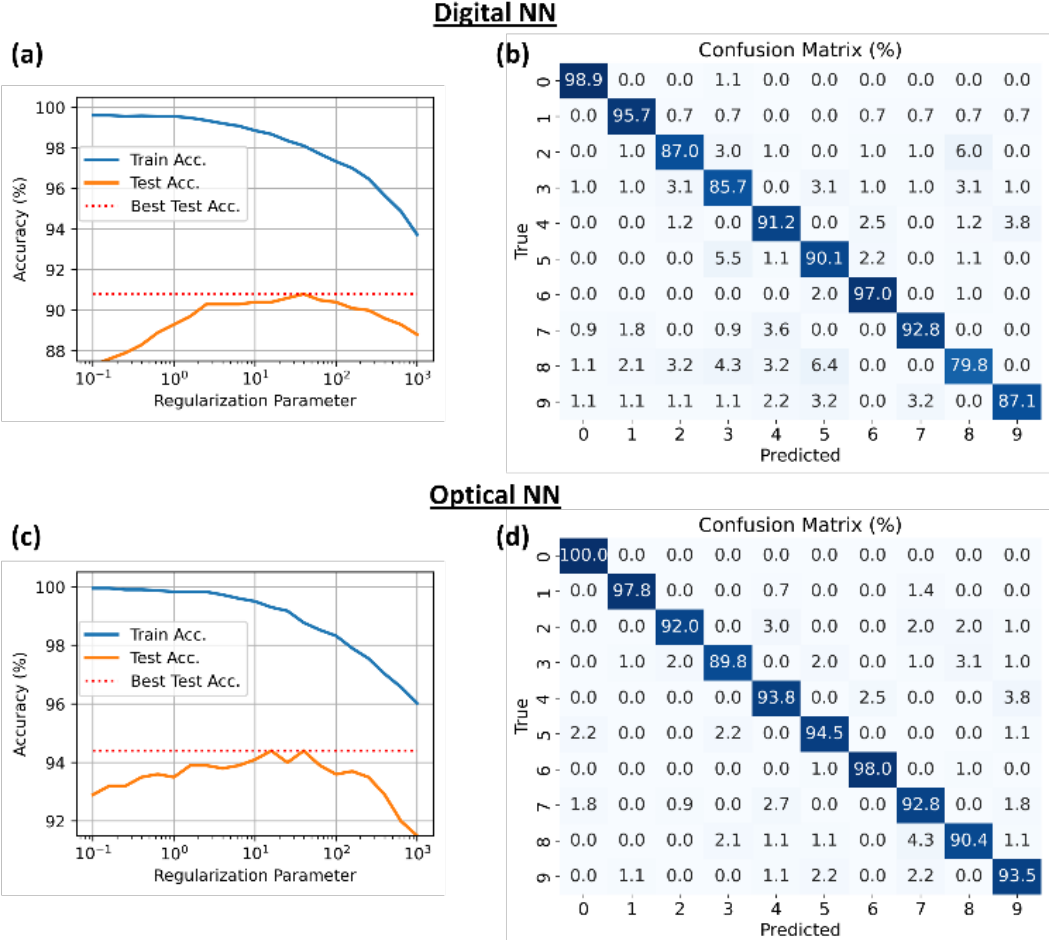

Fig. 3. Comparison between classification performances of NNs with and without optical transform on a subset of the MNIST-digits dataset (a) The dependence of training and test accuracies on the Ridge classifier regularization strength without the optical transform. (b) Confusion matrix of the test set when the optimum regularization parameter in (a) is used. (c) The dependence of training and test accuracies on the Ridge classifier regularization strength with the optical transform. (d) Confusion matrix of the test set when the optimum regularization parameter in (c) is used.

The first two trainable layers of the NN trained by EBP (Fig. 2a) and all trainable layers of FFA trained NN's (Fig. 2b and 2c) except their output layers are convolutional layers each with one trainable kernel sized 5x5 with 4 pixels of dilation ReLU nonlinearity. Dilated kernels capture large features in images with a small number of parameters and are suitable for the current implementation as speckles span across multiple pixels. Layer normalization operations are used as a part of the FFA and they scale activations for each sample so that the vector of activations has an L2 norm equal to 1. In the optical transform steps in Fig 2.c, these vectors of activations modulate beam phase distribution as 2-dimensional arrays individually, and the corresponding beam output patterns are recorded by the camera and transferred to the next trainable layer. The performances obtained on the test set are shown in Table 1 and they confirm the findings in [14], that the performance of a NN trained with the FFA decreases compared to the performance obtained with EBP trained network. For the given dataset, the digitally EBP trained NN (2a.) consisting of 2 convolutional and 1 FC layer achieves 91.8 % test accuracy while its FFA trained counterpart reaches 90.8 % test accuracy. On the other hand, the addition of the high-dimensional nonlinear mapping improves the performance of the 2 convolutional layer NN to 94.4% without any increase in trainable weights. Also, the fact that LeNet-5 [17] achieves only 95.0% test accuracy when trained with EBP on the same subset of MNIST digits (instead of ~1% error rate on the full dataset of 60000 samples) shows the potential of our proposed approach to reach higher accuracies with larger datasets.

Table 1 Comparison of Accuracies between Different Neural Networks

| Network | Test Accuracy (%) | Number of Parameters | Digital Operations per Sample (FLOPs) |
| --- | --- | --- | --- |
| 2 conv. + 1 FC - EBP | 91.8 | 14,398 | 143 K |
| 3 conv. + 1 FC - FFA | 90.8 | 26,712 | 204 K |
| 2 conv.+ optics + 1 FC – FFA | 94.4 | 24,638 | 150 K |
| LeNet – 5 - EBP | 95.0 | 61,706 | 846 K |

The improvement in the performance of the NN with the addition of optical transforms is shown in Figure 3 in more detail. The accuracy obtained is plotted as a function of the strength of the regularization term used in the training. The optical NN performs better with a stronger regularization, indicating a higher number of effective features provided by the optical transform. When the optical nonlinear connectivity is combined with the relatively small number of trainable weights in convolutional layers, both the training and test accuracy on the subset of the MNIST digits improve (Fig.3c). This improvement is observed also in class-wise accuracies in the confusion matrix. The correct inference ratio increases for nearly all classes with the optical transform.

Even though FFA simplifies NN training and decreases the memory usage by decoupling weight updates of different layers, benchmarks show that the task performance tends to decrease compared to training the same architecture with EBP. With this study, we demonstrate that by adding non-trainable nonlinear mappings to the architecture, this decrease can be reversed and even an increase in the performance can be obtained.

To make use of optical or any other type of physical transforms with a gradient-based or metaheuristic training algorithm, the transform should be applied to each sample multiple times through the epochs or iterations of the algorithm. FFA makes physical systems more accessible to NNs by removing this requirement. With the implementation presented in this letter, the physical transform is applied to the data representation only once after training each layer, and the next layer is trained with the transformed representation. This advancement could be a solution to one of the biggest bottlenecks in training optical NNs considering the limited modulation speed of electro-optic conversion devices.